\pdfoutput=1

\documentclass[11pt]{article}

\usepackage{EMNLP2023}

\usepackage{times}
\usepackage{latexsym}

\usepackage[T1]{fontenc}

\usepackage[utf8]{inputenc}

\usepackage{microtype}

\usepackage{inconsolata}
\usepackage{graphicx}

\title{In-context Interference in Chat-based Large Language Models}
\author{First Author \\
  Affiliation / Address line 1 \\
  Affiliation / Address line 2 \\
  Affiliation / Address line 3 \\
  \texttt{email@domain} \\\And
  Second Author \\
  Affiliation / Address line 1 \\
  Affiliation / Address line 2 \\
  Affiliation / Address line 3 \\
  \texttt{email@domain} \\}

\begin{document}
\maketitle
\begin{abstract}

Large language models (LLMs) have had a huge impact on society due to their impressive capabilities and vast knowledge of the world. Various applications and tools have been created that allow users to interact with these models in a black-box scenario.
However, one limitation of this scenario is that users cannot modify the internal knowledge of the model, and the only way to add or modify internal knowledge is by explicitly mentioning it to the model during the current interaction. This learning process is called in-context training, and it refers to training that is confined to the user’s current session or context.
In-context learning has significant applications, but also has limitations that are seldom studied. In this paper, we present a study that shows how the model can suffer from interference between information that continually flows in the context, causing it to forget previously learned knowledge, which can reduce the model's performance.
Along with showing the problem, we propose an evaluation benchmark based on the bAbI dataset.

\end{abstract}


\section{Introduction}

Chat-based Large Language Models (LLMs) \cite{devlin2019bert, openai2023gpt} have gained significant attention in the last year due to their impressive capabilities and potential to perform a wide range of tasks \cite{brown2020language}. These models have been used in various contexts, and multiple experts have been surprised by their remarkable ability to maintain a fluid conversation, answering questions with information stored in their weights and with information acquired within each session.

Despite their ability to hold conversations, the inability to modify model weights in many applications makes the only way to add relevant information is through prompts in the same context \cite{brown2020language, dai2022can}. The property to learn and accumulate knowledge of the model without the need to modify the weights is a critical tool for current and future LLMs that have not been thoroughly examined or studied \cite{wu2023openicl}.

Therefore, in this paper, we take a step toward understanding the limitations and strengths of in-context learning of chat-based LLMs by studying how the model behaves when we continually add new knowledge. These understandings can provide insights into how we can mitigate possible limitations and further improve the performance of these models. This will ensure reliable and efficient interactions with users.

The main contributions of this paper can be summarized as follows: First, we propose a benchmark to evaluate the accumulation and retention of information of LLMs, mainly to understand their capabilities and limitations of in-context learning. Second, using the proposed benchmark, we show evidence that these models can suffer from interference inside the same chat session, decreasing the performance as we add new information. This is critical, as for some applications, it may not be relevant to forget previous information within a session. However, there are cases where it is crucial that the model provides reliable information. Third, we provide insights into the current ability of these models to learn, retain, and reason on in-context scenarios. The problems presented in this work will facilitate the development of more robust and efficient language models and will pave the way for their successful integration into various applications and domains.

\section{Related Works}
In this section, we make a brief analysis of the different research areas that intersect with our work.

\textbf{In-Context Learning}
Recent papers explore in-context learning in chat-based LLMs, showing the importance of this training process \cite{garg2023transformers}. \citet{codaforno2023metaincontext} presents meta-in-context learning, showing recursive improvement of in-context learning on various tasks while using meta-learning. Others have compared the generalization of few-shot fine-tuning and in-context learning, emphasizing the role of model size and examples \cite{mosbach2023fewshot}. 

\textbf{Question Answering in LLMs}
Question answering is a key application of LLMs. Recent research has advanced this area with novel methods, like LAMOC, which uses language model-guided captioning for knowledge based visual question answering \cite{du2023zeroshot}. The McL-KBQA framework leverages in-context learning for knowledge base question answering in low-resource settings \cite{tan2023make}. The TempReason dataset measures and improves LLMs’ temporal reasoning capability \cite{tan2023benchmarking}. ChatGPT’s performance on complex questions reveals the challenges of long-term reasoning for chat-based LLMs \cite{tan2023evaluation}.

\textbf{Continual Learning in LLMs}
Adding new knowledge continually without forgetting is the focus of Continual Learning (CL). Most works in CL for LLMs focus on training soft prompts \cite{wang2022preserving, razdaibiedina2023progressive} or adapters \cite{ke2023continual}. These works are mainly focused on training the model weights to specific tasks and problems, unlike ours, which seeks to study how in-context learning influences the performance, that is, without modifying model weights. Others have studied the behavior of different methods to continuously learn \cite{araujo2022relevant, fu2023chainofthought}.



\section{In-Session evaluation}

Normally, users interact with chat-based LLM in terms of sessions or contexts. If we assume no access to internet, each session is a close environments where the user can interact with a model as a black-box, meaning the the user has no access to directly change or condition the weight of the model, and the only way to of interaction is through an input known as prompts via the chat interface. Ideally, when users interact with these models, they expect that the model remembers previous interaction inside the same context. This is know as in-context learning \cite{zhou2022teaching}, where the model learns through the information that the user provides in forms of input and users expect a perfect memory.

To test the capabilities and limitations to remember previous interactions, we propose to submit a sequence of stories or facts and test how the performance of the model evolves as the amount of information that stored increase. Starting from an input $s_0$, we ask questions $q_0$ associate to input $0$ to obtain performance $per_0$. At time step $i$, we will provide the model with an input $s_i$ and ask questions $q_{\leq i}$, meaning that we will ask all questions from input $0$ to $i$ to obtain performance $per_i$. 

If a model can correctly answer a question in a single context, we expect that the performance over time should not decrease as we add more information. However, as we will show in our experiments, the model suffers from interference and the performance decreases as we increase the number of facts.

\subsection{Benchmark}

To evaluate in-context learning, we need a sequence of inputs that the model should be able to solve quickly  without prior knowledge. For this reason, we need to evaluate only the retention capacity of the information delivered in the context.

Starting from a subset of the bAbI dataset \cite{DBLP:journals/corr/WestonBCM15}, we create a sequence of stories that give the model new information. The only way the model can correctly answer the questions is by using the information extracted from inside the context.

\begin{figure}
\includegraphics[width=\columnwidth]{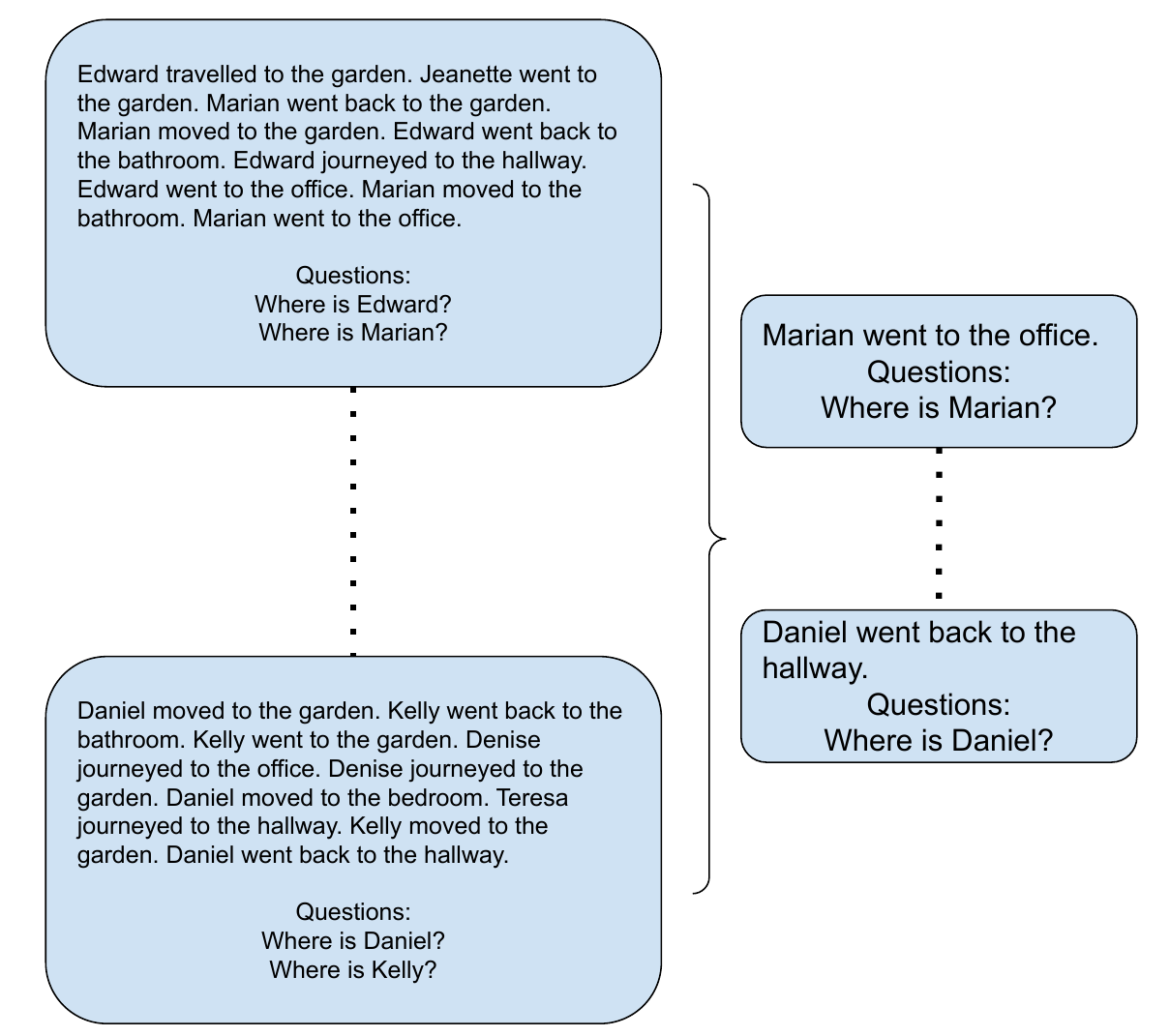} 
\caption{Modifications were made to the original stories to  simplify the problem. First, we replace all default names with unique names to minimize the risk of confusion. Second, we shortened the story, leaving only the last two sentences, removing the complexity of reasoning.}
\label{fig:sim_dataset}
\end{figure}

Given the complexity of the original dataset, we propose two modifications. In the first modification, we replaced all the default names of the entities in the dataset with unique names. By doing this, we ensured that the model could correctly identify the entities in the stories and answer questions about them, even when the entities were not explicitly named in the questions. In the second modification, we simplified the structure by including only the last two statements in each story, along with a single question about the last entity mentioned. This allowed us to assess whether the model could remember the context of the story and answer questions accurately, without the complexity of the original dataset. Figure \ref{fig:sim_dataset} provide an example of stories created from their original version.

A limitation of these models is the token limit. There are two things to take this into consideration, the first is the computational capacity required when we increase the number of tokens and the second is the size of the context that ensures effective performance of the model. To ensure this, we chose 50 stories that the model could correctly classify.

\subsection{Experimental setup}

To evaluate the previous scenario, we use the Vicuna model \cite{vicuna2023}, within the LangChain \cite{langchain23}framework. The main reason for using this model is that it is open source and this helps us to extend the experiments to other types of adaptation, as we will describe in future works.

Specifically, using the Hugging Face library \cite{wolf-etal-2020-transformers}, we load the \textbf{Vicuna13B} model, a LLaMa-based model for text generation. Due to computational limitations, we set the max number of tokens to 2048. Because the answer we are looking are only 1 word long, to reduce the probability of longer answers, we empirically found that a temperature of $0.7$ provide overall the best performance. Following previous work of in-context learning, we teach the model via a prompt that can be found in the Appendix.

\section{Experiments}

\subsection{One Context-One Story}

As mentioned in the previous section, to correctly detect if a model suffers from interference between new and old knowledge, it is important to detect when the performance decreases. For this, we must first verify that the model can correctly answer all questions in a one-story-per-context scenario to avoid noisy results that can affect the conclusions. For this, we first perform experiments with only one story and the corresponding questions per context. We reset the prompts after the interaction. The accuracy when using the full story is 58\%, as shown in Table \ref{tab:story_info}. This low accuracy and the high number of tokens per story encourage us to propose a simplified version of the known dataset. When testing the performance of the simplified version, the model obtained a 100\%, probing that the complexity of the original dataset could interfere with the conclusions drawn.

\begin{table}
\centering
\begin{tabular}{lcc}
\hline
 & \textbf{Full Story} & \textbf{Short Story}\\
\hline
Accuracy        & 58\% & 100\% \\
Avg. \# Tokens   & 67   & 19 \\
\end{tabular}
\caption{Accuracy and average number of token for both benchmarks. As expected, the complexity of the full benchmark makes the accuracy lower. However, when we simplify it, we can obtain a higher accuracy.}
\label{tab:story_info}
\end{table}

\subsection{Incremental stories}

When learning continually, one can expect that the model will accumulate all the information without forgetting previous interactions. This means that if the model correctly answers a question, and when new information is added, the model forgets the previous answer, we can identify an interference problem. Our hypothesis is that when new information is delivered, an interference problem occurs that confuses or erases some facts. Understanding this limitation would allow us to study the phenomenon and propose methods to accumulate information clearly. This would avoid the need to train the model and limit the number of tokens.

\begin{figure}
\includegraphics[width=\columnwidth]{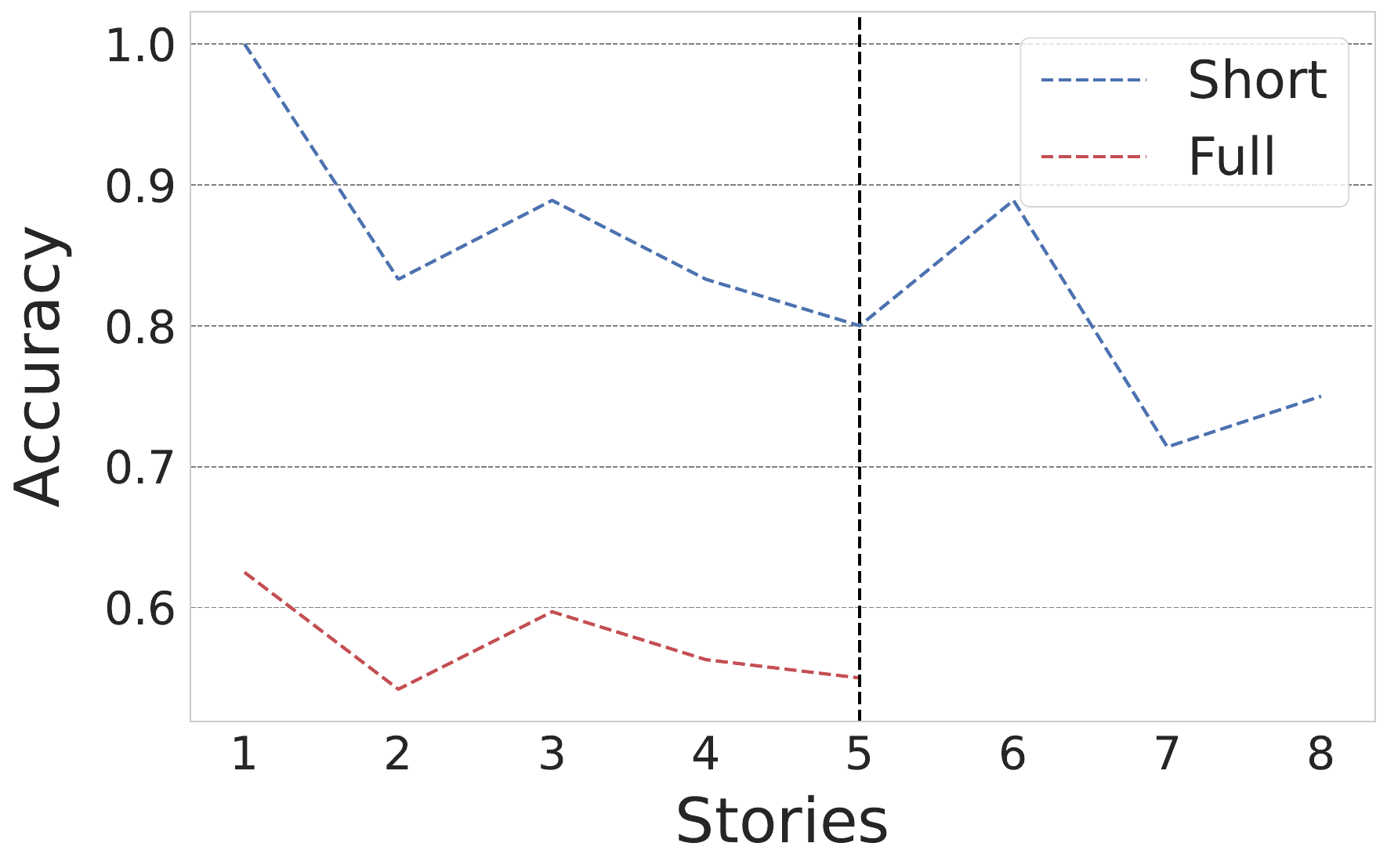} 
\caption{Evolution of the accuracy as we increase the number of stories in the context. As expected, as we increase the information in the context, we observed a decrease in the performance of the model.}
\label{fig:acc_accum}
\end{figure}

As shown in Figure \ref{fig:acc_accum}, as we add new knowledge to the prompts of the model, we can see that the performance of the model decrease from a 100\% with only one story to around 75\% when we have 8 in the same context. A similar effect appears when using the original stories, where the model show a decrease in the performance . It is important to note that we cannot  significantly increase  the amount of stories since we have a limitations on the amount of tokens the model can receive. Some studies have shown that is possible to increase the number of tokens \cite{bulatov2023scaling}, however, this normally increases the computational cost, as shown in Table \ref{tab:time_stories}, where we see that as we increase the size of the prompt ($\#$ of stories and length), the cost of delivering a response increases.

The interference between old and new information is not something new. As we train the weights of Deep Learning models, the modification of the weights cause a problem known as Catastrophic Forgetting (CF) \cite{mccloskey1989catastrophic}, and it is related to the constant modification of the weights of the model when we train new tasks. However, it is a different process than the one presented here where the model's weights are not modified. The interference in this case is at the information level and it is the model that is not able to correctly identify the relevant parts of all the information delivered.

\begin{table}
\centering
\begin{tabular}{lccccc}
\hline
\# Stories & \textbf{1} & \textbf{2} & \textbf{3} & \textbf{4} & \textbf{5}\\
\hline
Short & 34s & 38s & 44s & 58s & 63s \\
Full  & 46s & 49s & 71s & 78s & 85s \\
\end{tabular}
\caption{Time in seconds that it take to generate an answer as we increase the number of stories. By using smaller simplify version of the stories, we reduce the number of tokens which translates into a decrease in response time.}
\label{tab:time_stories}
\end{table}

\subsubsection{Summarizing}

\begin{figure}
\includegraphics[width=\columnwidth]{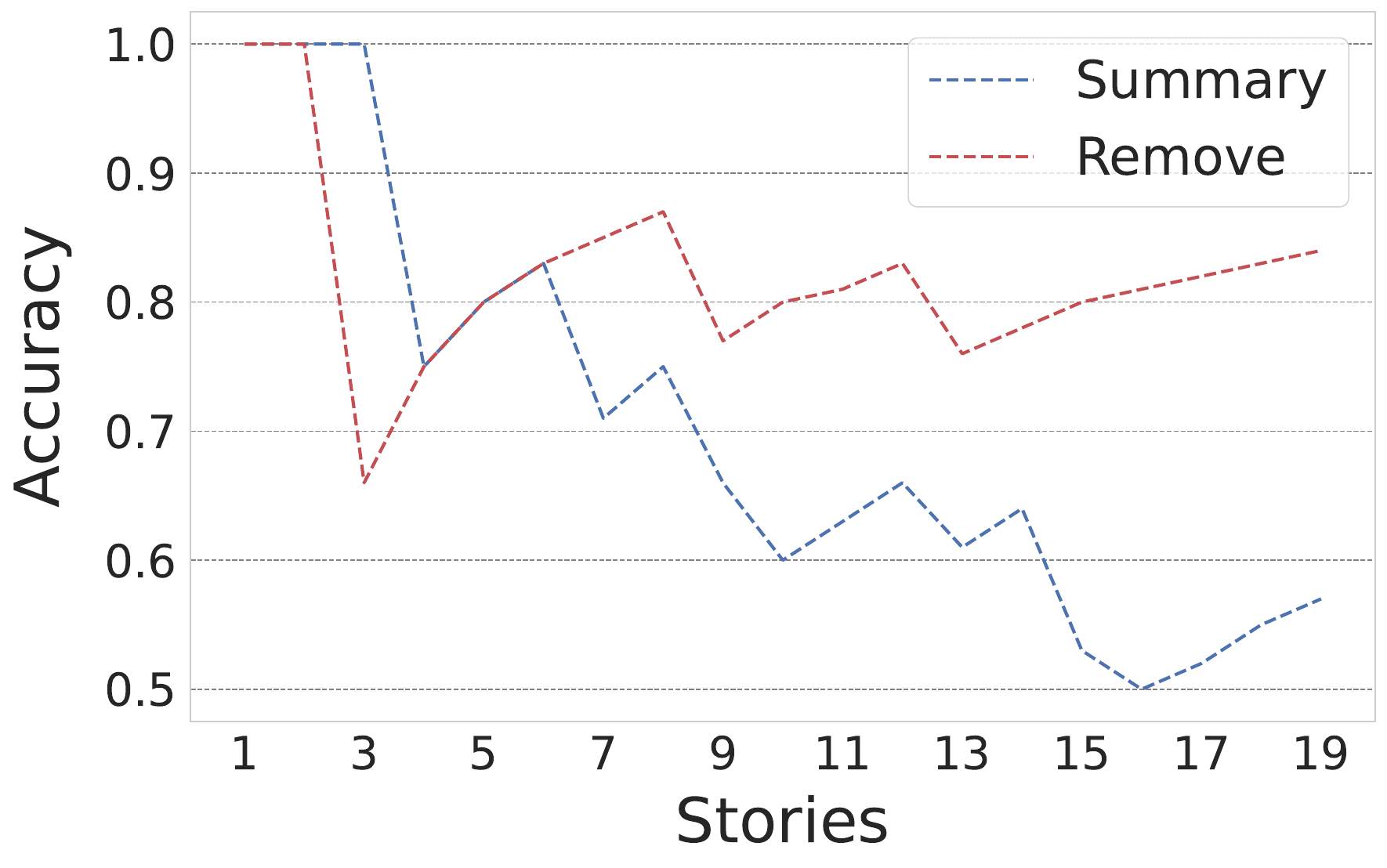} 
\caption{To mitigate the interference, we propose summarizing or removing previous information. By summarizing the previous story, we expect the model to keep only relevant information, however we observed a constant decrease in the performance. On the other hand, by removing old stories and keeping only 6 in the buffer, the model is able to reduce the interference.}
\label{fig:acc_summ}
\end{figure}

Similar to the CF problem, we need to devise ways to minimize interference between the information the model accumulates. One way is to apply summarizing tools that these methods have built-in, this can reduce the information but keep relevant knowledge in the buffer. By keeping what is strictly necessary, the model can compress the information, reducing the interference with unnecessary information and improve performance. However, as we can observed in Figure \ref{fig:acc_summ}, when summarizing the information the model do not maintain the performance.  

One theory of the above is that the model may be interfering with the responses it has delivered (which are part of the context). For this reason, instead of summarizing the previous information, we decided to delete old stories and only keep in buffer $6$ of them. By removing old stories, we are removing the option for the model to change its response (for better or worse), but more importantly we are reducing the amount of tokens that cause interference. Figure \ref{fig:acc_summ} shows how the model is able to keep the performance even with a high number of stories.

\section{Conclusion}

This paper introduces a new method to evaluate how chat-based LLMs are affected by interference in in-context learning. We propose a benchmark based on the bAbI dataset to test how well the models can accumulate information in the context. We find that adding new information causes interference that can harm the performance of the models in some scenarios where the user has no control over them. This is a preliminary study of the limitations of in-context learning, but it can be expanded to other related issues, such as the interference between the context-based knowledge and the pre-trained knowledge of the models.

\section*{Limitations}

Despite the efforts put into carrying out the experiments, it is important to note that the results are obtained from a single model (Vicuna 13B). Although these experiments can also be carried out on private models, it is important to highlight that in order to continue this line of research, it is necessary to have complete availability of the models, such as to study how the accumulation of in-context learning can affect previously acquired knowledge stored in the weights or adapters of a model.

\section{Acknowledgements}
This research was supported by Leonardo Labs. We acknowledge and appreciate them  for their valuable feedback and guidance throughout this project.



\bibliography{emnlp_subm/reference}
\bibliographystyle{acl_natbib}

\appendix

\section{Prompt}
\label{sec:appendix}

You will be provided with stories consisting of sequential actions performed by different characters, follow the actions of each character to the end.
HINT: THE FIRST POSITION OR ACTION CAN BE IGNORED IF THAT IS NOT THE ONLY PLACE A CHARACTER VISITED. IF A CHARACTER IN THE STORY MOVES TO 3 PLACES , FOLLOW THOSE ACTIONS ADD REPORT ONLY THE LAST VISITED OR JOURNEYED LOCATION.
Keep reasoning to a minimum and just follow the actions sequentially to determine the final position of a character.
verbs like 'journeyed to','went to','traveled to','moved to','went back to'  mean 'is in'. for example, 'john traveled to the bedroom' means 'john is in the bedroom'.
Below is an example story and corresponding questions:

Story:
Mario moved to the school. Kyle went to the cafeteria. Nathan went back to the cafeteria. Tanya moved to the library. Kyle moved to the school. Tanya journeyed to the school. Mario moved to the cafeteria. Nathan travelled to the school. Kyle went back to the library. Kyle moved to the bedroom.
Questions:
Where is Kyle?
Where is Tanya?

Answer:
Bedroom
School

NOTE: KEEP THE ANSWERS SHORT , AS SEEN IN THIS EXAMPLE
You memory will be tested on Stories you see overtime. Try as much as possible to keep the stories in memory and also the answers you give.KEEP ANSWERS CONSISTENT.( Since we want to eveluate how much you forget)

\end{document}